%% file: main.tex
\documentclass{article}

\usepackage{xcolor} 
\usepackage[preprint]{corl_2026} 

\input{setup/packages}
\input{setup/math}

\input{setup/defs}

\title{NAC: Neural Action Codec for Vision-Language-Action Models}

\author{
  Ahad Jawaid \quad
  Yu Xiang \\
  \\
  The University of Texas at Dallas  \\
}

\hypersetup{
  pdftitle={NAC: Neural Action Codec for Vision-Language-Action Models},
  pdfauthor={Ahad Jawaid and Yu Xiang}
}

\begin{document}

\maketitle


\input{content/abstract}



\input{content/body}


\clearpage


\bibliography{main}  

\appendix
\section*{Appendix}
\input{content/appendix}

\end{document}

%% file: setup/packages.tex
\usepackage{algorithm}
\usepackage{algorithmicx}
\usepackage{algcompatible}
\usepackage[noEnd,commentColor=black]{algpseudocodex}

\usepackage{amsmath}
\usepackage{amssymb}
\usepackage{mathtools}
\usepackage{amsthm}
\usepackage{cancel}

\usepackage{thmtools} 
\usepackage{thm-restate}

\usepackage{hyperref}       
\usepackage[capitalize,noabbrev]{cleveref}

\crefname{assumption}{Assumption}{Assumption}
\theoremstyle{plain}

\usepackage[textsize=tiny]{todonotes}

\usepackage[utf8]{inputenc} 
\usepackage[T1]{fontenc}    

\usepackage{url}            
\usepackage{booktabs}       

\usepackage{times}

\usepackage{xcolor} 
\usepackage{graphicx}
\usepackage{blindtext}
\usepackage[labelformat=empty, position=top]{subcaption}
\usepackage{etoolbox}
\makeatletter
\renewenvironment{table}{%
  \setlength{\abovecaptionskip}{4pt}%
  \setlength{\belowcaptionskip}{2pt}%
  \@float{table}}{\end@float}
\makeatother
\AtBeginEnvironment{table*}{%
  \setlength{\abovecaptionskip}{4pt}%
  \setlength{\belowcaptionskip}{2pt}%
}
\usepackage[export]{adjustbox}

\usepackage{wrapfig}
\usepackage{pifont}

\usepackage{amsfonts}       
\usepackage{nicefrac}       
\usepackage[nopatch=showhyphens]{microtype} 

\usepackage{multirow}
\usepackage{multicol}
\usepackage{soul}
\usepackage{tikz-cd}

\usepackage{transparent}

\usepackage{tcolorbox}
\tcbuselibrary{theorems}

\definecolor{ourcolor}{HTML}{00A89D}
\definecolor{ourpurple}{HTML}{CB297B}
\definecolor{ourblue}{HTML}{0076BA}
\definecolor{ourmiddle}{HTML}{66509B}
\definecolor{ourlightblue}{HTML}{F5FAFE}
\definecolor{ourlightmiddle}{HTML}{F7F6F9}

\declaretheoremstyle[
    headfont=\bfseries,
    bodyfont=\normalfont,
    spaceabove=10pt, spacebelow=10pt,
    headpunct={},
    postheadspace=1em,
    mdframed={
        backgroundcolor=ourcolor!5,
        linecolor=ourcolor!50!black,
        linewidth=1pt,
        roundcorner=4pt
    }
]{colorboxed}

\declaretheoremstyle[
    headfont=\bfseries,
    bodyfont=\normalfont,
    spaceabove=10pt, spacebelow=10pt,
    headpunct={},
    postheadspace=1em,
    mdframed={
        backgroundcolor=ourblue!5,
        linecolor=ourblue!50!black,
        linewidth=1pt,
        roundcorner=4pt
    }
]{bluecolorboxed}

\declaretheoremstyle[
    headfont=\bfseries,
    bodyfont=\normalfont,
    spaceabove=10pt, spacebelow=10pt,
    headpunct={},
    postheadspace=1em,
    mdframed={
        backgroundcolor=ourpurple!5,
        linecolor=ourpurple!50!black,
        linewidth=1pt,
        roundcorner=4pt
    }
]{purplecolorboxed}

\declaretheoremstyle[
    headfont=\bfseries,
    bodyfont=\normalfont,
    spaceabove=10pt, spacebelow=10pt,
    headpunct={},
    postheadspace=1em,
    mdframed={
        backgroundcolor=black!5,
        linecolor=black!50!black,
        linewidth=1pt,
        roundcorner=4pt
    }
]{greycolorboxed}

\usepackage{tikz}
\usetikzlibrary{decorations.pathreplacing,calc}

\newcommand{\ourshort}{NAC}

%% file: setup/math.tex
\usepackage{amsmath,amsfonts,bm, stackengine}

\DeclarePairedDelimiterX{\infdivx}[2]{(}{)}{%
  #1\;\delimsize\|\;#2%
}

\def\eqref#1{equation~\ref{#1}}

\def\1{\bm{1}}

\DeclareMathAlphabet{\mathsfit}{\encodingdefault}{\sfdefault}{m}{sl}
\SetMathAlphabet{\mathsfit}{bold}{\encodingdefault}{\sfdefault}{bx}{n}



%% file: setup/defs.tex
\definecolor{darkblue}{rgb}{0,0.08,0.45}
\definecolor{cred}{HTML}{D62728}
\definecolor{cblue}{HTML}{1F77B4}
\definecolor{cgreen}{HTML}{79AB76}
\definecolor{cgrey}{rgb}{0.6,0.6,0.6}

\renewcommand{\mathbf}{\boldsymbol}

\makeatletter

%% file: content/abstract.tex
\begin{abstract}
Vision-language-action (VLA) models rely on discrete action tokenizers to bridge continuous robot control and autoregressive sequence modeling, yet existing tokenizers often trade off between compression, latency, and downstream performance. We revisit this design through the lens of neural audio codecs—convolutional encoder–decoder architectures with residual vector quantization that serve as the standard front end for audio foundation models. Motivated by their success, we introduce the Neural Action Codec (NAC), which treats short robot action trajectories as multi-channel 1D signals and compresses them using a multi-scale RVQGAN architecture. With adaptations to the action representation, compression rate, and reconstruction objective, audio-codec-style models can autoencode actions with high fidelity without substantial architectural changes. NAC provides a compact, ordered token space via offset codebooks, enabling standard autoregressive policies to operate over short, structured sequences. Meanwhile, a Vocos-style decoder with an ISTFT head and adversarial discriminators recovers action trajectories. Across LIBERO-10, RoboMimic, and a suite of real-world manipulation tasks, NAC achieves high reconstruction fidelity and higher average success rates than binning, FAST, and prior VQ-based tokenizers at comparable or better compression rates. These results demonstrate that repurposed neural audio codecs offer a strong, practical backbone for learned action tokenization in modern VLAs.
\end{abstract}

%% file: content/body.tex
\begin{figure}[ht]
    \centering
    \includegraphics[width=0.8\linewidth]{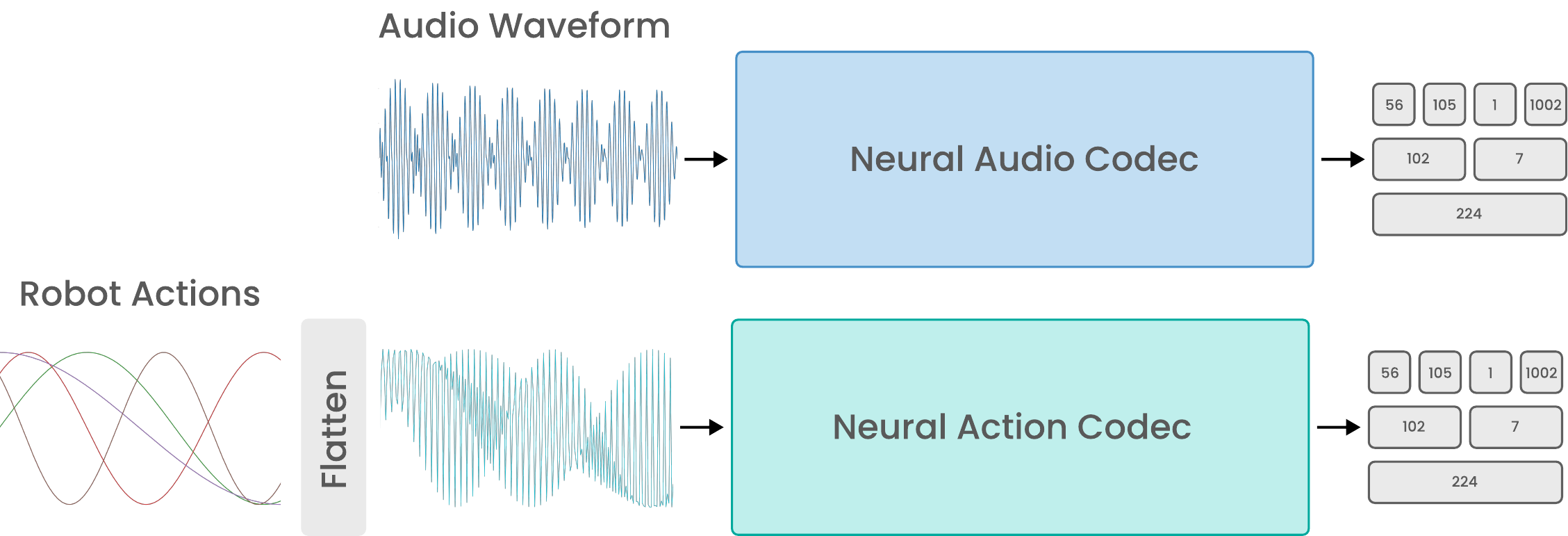}
    \caption{Neural audio codecs \citep{zeghidour2021soundstream, defossez2023encodec, kumar2023dac} adapted for action tokenization. \textbf{Top:} Modern neural codecs compress raw waveforms into compact, multi-scale discrete codes, preserving coarse structures and fine temporal details. \textbf{Bottom:} NAC applies this approach to robot action chunks, treating actions as multi-channel 1D signals to learn a highly compressed, discrete latent space for downstream autoregressive policy learning.}
    \label{fig:teaser}
\end{figure}

\section{Introduction}

Large-scale vision-language-action (VLA) models~\citep{black2024pi0,black2025pi0.5} map visual observations and natural language instructions directly to low-level control signals. Formulating robot control as a sequence modeling task \citep{openvla2024, liu2026oat} allows VLAs to inherit the generalization and reasoning capabilities of pre-trained vision-language models (VLMs) \citep{yang2025qwen3}. However, training VLAs requires action tokenization \citep{openvla2024, pertsch2025fast} to represent continuous physical actions as discrete symbols.

The action tokenizer is a critical design choice in VLA pretraining. Early methods relied on uniform per-dimension binning \citep{openvla2024}, producing prohibitively long token sequences for high-frequency control. Subsequent approaches, like FAST \citep{pertsch2025fast, lee2025molmoact}, used frequency-domain compression to shorten sequences. This improved performance, demonstrating that compressing the action space simplifies the generative modeling task for the VLA \citep{pertsch2025fast}.

Despite these advances, designing an optimal action tokenizer remains challenging \citep{liu2026oat}. The primary difficulty lies in capturing statistical regularities that enable the policy to model the underlying action distribution effectively. A secondary challenge is latency, dictated by the compression rate (tokens per action chunk) and the tokenizer's decoding speed.

The audio generation domain has extensively studied and largely mitigated similar challenges—high-fidelity compression, low latency, and complex temporal distributions. Neural audio codecs \citep{zeghidour2021soundstream, defossez2023encodec, kumar2023dac} use convolutional encoder-decoders with Residual Vector Quantization (RVQ) \citep{lee2022autoregressive} to compress waveforms into discrete codes. These scalable architectures form the foundation of modern audio foundation models \citep{borsos2023audiolm, wang2023valle}.

Audio and robotic actions share a continuous time-series structure but differ notably. Action sequences operate at lower frequencies (e.g., 30-60 Hz) \citep{cadene2026lerobot, pertsch2025fast} than audio (16-48 kHz) \citep{kaneko2022istftnet}. Actions are also multi-channel (typically 7-14 dimensions for robot joints or end-effectors) \citep{cadene2026lerobot}, whereas audio is primarily single or dual-channel. Despite these differences, the core objective of compressing continuous spatio-temporal signals remains fundamentally similar.

We propose the Neural Action Codec (NAC), which adapts state-of-the-art audio codec architectures for robotic action tokenization. Figure~\ref{fig:teaser} illustrates this transfer from audio codecs to action chunks. We retain the convolutional encoder--decoder and multi-scale RVQ design of neural audio codecs \citep{siuzdak2024snac, kumar2023dac}, adapting the action representation, temporal downsampling, and reconstruction objective for robot trajectories.

Our main contributions are as follows:
\begin{itemize}
    \item We show that neural audio codec techniques can be applied to robot actions through NAC, which adapts a multi-scale RVQGAN architecture \citep{siuzdak2024snac, kumar2023dac} for action tokenization.
    \item We describe how differences in sampling rate and channel structure guide the action representation and temporal downsampling, and analyze how reconstruction objectives, adversarial training, and decoder design affect performance.
    \item We formulate an autoregressive behavioral cloning policy that uses NAC's offset codebooks for structured, causal next-token prediction.
    \item We empirically validate our approach's reconstruction fidelity and downstream policy performance across simulated and real-world benchmarks.
\footnote{Visit the project page at \url{https://ahadjawaid.com/nac} or contact the author at \texttt{ahad.jawaid@utdallas.edu}.}
\end{itemize}

\section{Related Work}

\subsection{Audio Codecs and Audio Foundation Models}
Neural audio codecs \citep{zeghidour2021soundstream, defossez2023encodec, kumar2023dac} compress waveforms into discrete codes using convolutional encoders and RVQ. SoundStream \citep{zeghidour2021soundstream} paired SEANet encoders with RVQ for low-bitrate compression; DAC \citep{kumar2023dac} improved fidelity via quantization bottlenecks and discriminator design; and SNAC \citep{siuzdak2024snac} introduced multi-scale RVQ to operate at varied temporal resolutions. These discrete codes enable audio foundation models to treat generation as sequence modeling over codec tokens \citep{borsos2023audiolm, wang2023valle, yang2023uniaudio}. We adapt this codec recipe for robotic action sequences.

\subsection{Vision-Language-Action Models}
Vision-Language-Action (VLA) models extend pretrained vision-language models (VLMs) \citep{beyer2024paligemmaversatile3bvlm, yang2025qwen3} into generalist robot policies. A recurring bottleneck is mapping continuous action trajectories into a compact discrete form for efficient modeling. Autoregressive VLAs like OpenVLA \citep{openvla2024} fine-tune Llama-2 \citep{touvron2023llama2openfoundation} to predict discrete action tokens directly, making the tokenizer central to sequence length, compression, and learnability. This focus extends to broader foundation systems: MolmoACT \citep{lee2025molmoact, fang2026molmoact2actionreasoningmodels} adopts FAST tokenization \citep{pertsch2025fast}, and \(\pi_0\)-family models \citep{black2024pi0, black2025pi0.5, intelligence2026pi07steerablegeneralistrobotic} depend on effective action representations during training. This suggests that even models with continuous low-level action heads benefit from grounding in a strong discrete action space. NAC targets this setting by learning representations from neural codec principles rather than relying purely on hand-designed compression.

\subsection{Discrete Action Tokenizers}
Early VLAs used naive per-dimension binning \citep{openvla2024}, yielding prohibitively long sequences. FAST \citep{pertsch2025fast} compresses actions via the Discrete Cosine Transform (DCT) \citep{ahmed1974discrete} and Byte-Pair Encoding (BPE) \citep{sennrich2016neural}, but its hand-designed frequency prior may miss complex, non-linear dynamics. Learned tokenizers offer a data-driven alternative: VQ-VLA \citep{wang2025vqvla} uses vector quantization \citep{gray1984vector}; ActionCodec \citep{dong2026actioncodec} and FASTer \citep{liu2026faster} employ RVQ, with FASTer utilizing a transformer autoencoder \citep{vaswani2017attention} and DCT L1 reconstruction \citep{zhao2016loss}; and OAT \citep{liu2026oat} enforces causally ordered token spaces via nested dropout \citep{rippel2014learning} and register tokens \citep{darcet2024vision}.

\ourshort{} bridges state-of-the-art audio codecs \citep{siuzdak2023vocos, siuzdak2024snac} and action modeling. Unlike FASTer and OAT, it uses a fully convolutional SEANet encoder with multi-scale RVQ and adversarial training, optimizing for action reconstruction fidelity.

\section{Method}

In this section, we introduce the Neural Action Codec (\ourshort{}). \ourshort{} maps continuous action chunks to 1D signals, compresses them via a convolutional encoder and multi-scale RVQ \citep{siuzdak2024snac, lee2022autoregressive}, and decodes them via a Vocoder-style decoder. The decoder predicts spectral magnitude and phase, then reconstructs the action signal using an inverse short-time Fourier transform (ISTFT) \citep{kaneko2022istftnet}. We then detail the autoregressive behavioral cloning formulation \citep{chen2021decision, zhao2023learning} that leverages \ourshort{}'s structured token space.

\subsection{Preliminaries}

We formulate robotic control via behavioral cloning \citep{torabi2018behavioral}. Given a history of visual observations \(o_{\text{img}} \in \mathbb{R}^{H_{\text{o}} \times N_{\text{cam}} \times C \times H \times W}\) (where \(H_{\text{o}}\) is history length, \(N_{\text{cam}}\) is number of cameras, and \(C, H, W\) are image dimensions) and a task condition \(l\) (a natural language instruction or discrete task UID), the policy predicts a contiguous chunk of future actions \(a_{1:H_{\text{a}}} \in \mathbb{R}^{B \times H_{\text{a}} \times D_{\text{a}}}\). Here, \(B\) is batch size, \(H_{\text{a}}\) is action horizon, and \(D_{\text{a}}\) is action dimensionality.

Autoregressive sequence models require a tokenizer \(\mathcal{T}\) to map the continuous action chunk \(a_{1:H_{\text{a}}}\) into a discrete token sequence \(C = [c_1, \dots, c_L]\), where \(c_i \in \mathcal{V}\) for a discrete vocabulary \(\mathcal{V}\), and \(L\) is the total number of multi-scale tokens. Regressing continuous actions directly can suffer from compounding errors \citep{zhao2023learning} and struggles with multimodal action distributions \citep{openvla2024}. Mapping chunks into discrete tokens allows the policy \(\pi_\theta\) to leverage the expressive modeling capabilities of autoregressive transformers by maximizing the log-likelihood of the token sequence: 
\begin{equation}
\max_\theta \mathbb{E} \left[ \sum_{i=1}^{L} \log \pi_\theta(c_i \mid o_{\text{img}}, l, c_{<i}) \right],
\end{equation}
where \(c_{<i} = [c_1, \dots, c_{i-1}]\) denotes previously generated tokens.

During inference, the policy autoregressively generates a token sequence \(\hat{C}_{1:L}\). A detokenization mapping then recovers the executable continuous actions:
\begin{equation}
    \mathcal{T}^{-1}: \hat{C}_{1:L} \mapsto \hat{a}_{1:H_{\text{a}}}
\end{equation}

\begin{figure}[t]
    \centering
    \includegraphics[width=\linewidth]{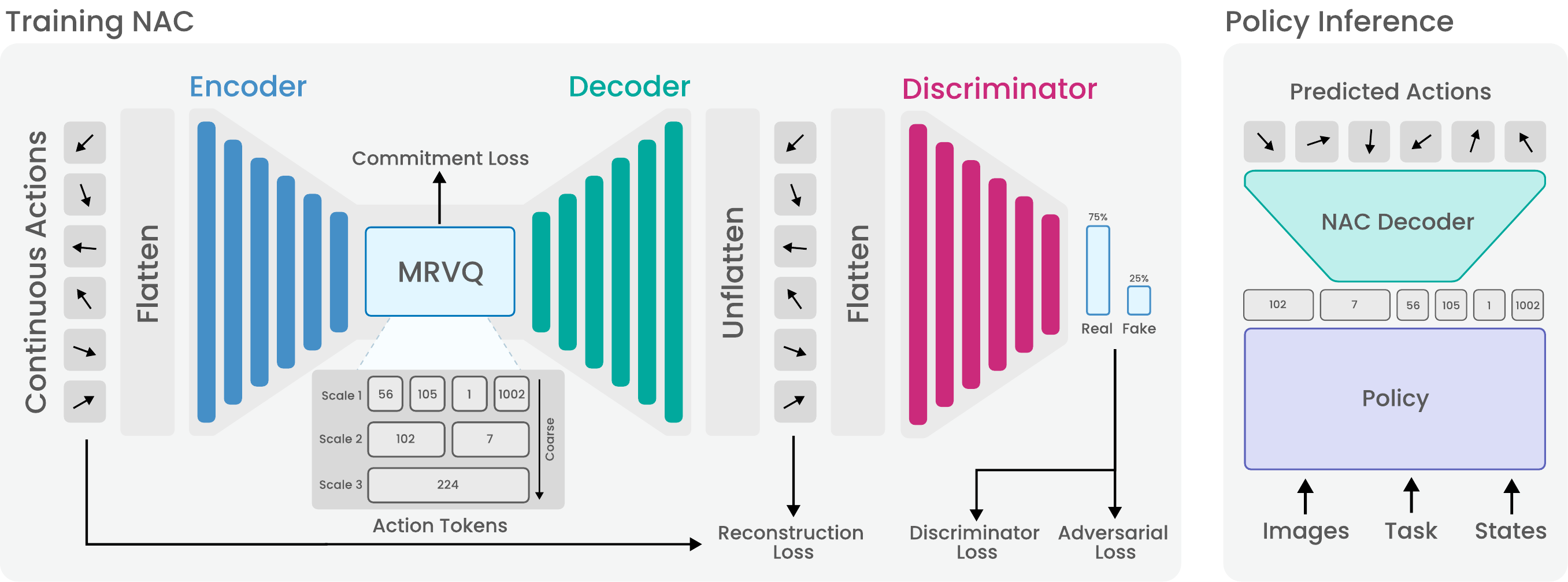}
    \caption{\textbf{NAC overview:} A continuous action chunk is flattened into a 1D pseudo-waveform and encoded by a SEANet-style encoder \citep{tagliasacchi20_interspeech}. Multi-scale residual vector quantization \citep{siuzdak2024snac} compresses the latent into discrete codes at progressively finer temporal resolutions. A Vocos-style decoder \citep{kong2020hifi, siuzdak2023vocos} with an ISTFT head reconstructs the action chunk. The policy then models the resulting offset code sequence autoregressively for downstream control.}
    \label{fig:architecture}
\end{figure}

\subsection{\ourshort{} Tokenizer}

\textbf{1D Signal Representation.} We treat the continuous action chunk \(a \in \mathbb{R}^{B \times H_{\text{a}} \times D_{\text{a}}}\) as a 1D multichannel signal. We first flatten the actions along the temporal and feature dimensions to form \(a_{\text{flat}} \in \mathbb{R}^{B \times (H_{\text{a}} \cdot D_{\text{a}})}\). This is unsqueezed into a single-channel 1D pseudo-waveform \(w \in \mathbb{R}^{B \times 1 \times L}\), where \(L = H_{\text{a}} \cdot D_{\text{a}}\). Optionally, per-dimension linear normalization is applied based on dataset statistics prior to flattening. Flattening reuses the single-channel audio architecture, with sample position identifying timestep and action dimension.

\textbf{SEANet Encoder.} The pseudo-waveform \(w\) is processed by a fully convolutional SEANet-style encoder \citep{zeghidour2021soundstream}. It consists of stacked 1D convolutions \citep{kiranyaz20211d} with strided downsampling defined by ratios \(R = [r_1, \dots, r_k]\). The total temporal downsampling factor (hop length) is \(\prod R\). We reduce it relative to the audio configuration to suit action token rates. The network uses residual blocks \citep{he2016deep} with ELU activations \citep{clevert2015fast}, weight normalization \citep{salimans2016weight}, and reflection padding \citep{isola2017image}. The output is a continuous latent representation \(z \in \mathbb{R}^{B \times D_{\text{enc}} \times T^\prime}\), where \(D_{\text{enc}}\) is the latent dimension (e.g., 512) and \(T^\prime = L / \prod R\). Figure~\ref{fig:architecture} overviews the full tokenizer and policy pipeline.

\textbf{Multi-Scale Residual Vector Quantization (MRVQ).} To discretize \(z\), we apply Multi-Scale RVQ \citep{siuzdak2024snac} over \(n_{\text{q}}\) quantization stages, each with a codebook of size \(V_{\text{bins}}\) (e.g., 1024). Codebook sizes and pooling scales follow the audio recipe. Unlike standard RVQ \citep{lee2022autoregressive}, we apply per-stage temporal pooling prior to quantization. This forces earlier stages to compress information over wider temporal windows to capture coarse, global trajectory structures, while later stages capture fine-grained, high-frequency residuals.

Let the unquantized residual at stage \(s\) be \(z_s\) (with \(z_0 = z\)). The quantized output is the nearest neighbor in the codebook:
\begin{equation}
    c_s = \arg\min_{j \in V_{\text{bins}}} || \text{pool}(z_s) - e_j^{(s)} ||_2
\end{equation}
where \(\text{pool}(\cdot)\) is a temporal average pooling operation, and \(e_j^{(s)}\) is the \(j\)-th learned embedding in the \(s\)-th codebook. Codebooks are trained jointly with the tokenizer: k-means initialization on the first batch \citep{macqueen1967multivariate}, then exponential moving average (EMA) updates at every forward pass, keeping each codebook vector as a slow running average of the encoder features assigned to that code. We upsample embeddings back to the encoder resolution and subtract them to form the residual for the next stage: \(z_{s+1} = z_s - \text{upsample}(e_{c_s}^{(s)})\). The final quantized latent is the sum of the upsampled embeddings.

\textbf{Commitment Loss.} We apply a \textit{commitment loss} \citep{van2017neural} to prevent the encoder's continuous outputs from arbitrarily diverging from the learned codebook embeddings. Without this regularization, the encoder could scale its outputs indefinitely, causing codebook collapse \citep{roy2018theory} where only a few vectors are used. The commitment loss penalizes the squared \(L_2\) distance between the unquantized latents and the chosen codebook vectors:
\begin{equation}
    \mathcal{L}_{\text{commit}} = \| z - \text{sg}(e_c) \|_2^2
\end{equation}
where \(\text{sg}(\cdot)\) denotes the stop-gradient operator, so codebook vectors are updated by EMA rather than through this loss. 

\textbf{Decoder and ISTFT Head.} The quantized latent passes to a Vocos-style backbone \citep{siuzdak2023vocos,siuzdak2024snac} consisting of a Conv1D embedding layer, ResNet \citep{targ2016resnet} blocks with attention, and a stack of ConvNeXt \citep{liu2022convnet} blocks. To map back to the 1D signal space, we employ an Inverse Short-Time Fourier Transform (ISTFT) Head \citep{kaneko2022istftnet}. The network predicts STFT magnitude and phase, and the ISTFT operation reconstructs the 1D signal. Crucially, the ISTFT hop length synchronizes with the encoder's hop length. This decoder functions as our detokenizer \(\mathcal{T}^{-1}\), projecting discrete latents back to continuous action space.

\textbf{Discriminator and Adversarial Loss.} Neural codecs use adversarial training via Generative Adversarial Networks \citep{goodfellow2014generative} to improve high-frequency fidelity. Audio models employ Multi-Period Discriminators (MPD) \citep{kong2020hifi} for periodic patterns, and Multi-Resolution Discriminators (MRD) \citep{jang2021univnet} or DAC discriminators \citep{kumar2023dac} for wide-band frequency and phase structures. Following the SNAC architecture \citep{siuzdak2024snac}, we adapt these discriminators for 1D action signals. We use the DAC discriminator in NAC and compare it with MPD, MRD, and training without a discriminator in Section~\ref{sec:codec_ablations}.

\textbf{Reconstruction Loss.} We use mean squared error (MSE) to reconstruct the flattened action signal. We evaluate alternative reconstruction losses in Section~\ref{sec:codec_ablations}. We formulate our generator loss \(\mathcal{L}_{\text{gen}}\) as:
\begin{equation}
    \mathcal{L}_{\text{gen}} = \lambda_{\text{rec}} \mathcal{L}_{\text{reconst}} + \lambda_{\text{commit}} \mathcal{L}_{\text{commit}} + \lambda_{\text{adv}} \mathcal{L}_{\text{adv}},
\end{equation}
Here, \(\mathcal{L}_{\text{reconst}}\) is the reconstruction loss and \(\mathcal{L}_{\text{adv}}\) is the adversarial loss from the DAC discriminator. The loss weights are given in Appendix Table~\ref{tab:tok_hparams}. The overall training procedure is summarized in Algorithm~\ref{algo:train_our_tok}.

\begin{algorithm}[tb]
\caption{\ourshort{} Tokenizer Training}
\label{algo:train_our_tok}
\begin{algorithmic}[1]
\Require Dataset $\mathcal{D}$ of action chunks $\{a_{1:H_{\text{a}}}\}$; SEANet encoder $E_\phi(\cdot)$; Multi-Scale RVQ $\mathcal{Q}(\cdot)$; Vocos-style decoder $D_\theta(\cdot)$; Discriminator $D_{\text{adv}}(\cdot)$.
\State Initialize codebook embeddings via k-means on the first training batch
\While{not converged}
    \State Sample action chunk $a_{1:H_{\text{a}}} \sim \mathcal{D}$
    \State Flattening: $w \gets \text{Flatten}(a_{1:H_{\text{a}}}) \in \mathbb{R}^{B \times 1 \times (H_{\text{a}} \cdot D_{\text{a}})}$
    \State Encoding: $z \gets E_\phi(w)$
    \State Quantization: $\hat{z}, C_{1:L_{\text{c}}}, \mathcal{L}_{\text{commit}} \gets \mathcal{Q}(z)$ \Comment{EMA-update codebooks}
    \State Decoding: $\hat{w} \gets D_\theta(\hat{z})$
    \State Reconstruction loss: $\mathcal{L}_{\text{reconst}} \gets \text{MSE}(w, \hat{w})$ \Comment{Or some other reconstruction loss}
    \State Generator loss: $\mathcal{L}_{\text{gen}} \gets \lambda_{\text{rec}} \mathcal{L}_{\text{reconst}} + \lambda_{\text{commit}} \mathcal{L}_{\text{commit}} + \lambda_{\text{adv}} \mathcal{L}_{\text{adv}}(\hat{w}, D_{\text{adv}})$
    \State Generator opt: $\{ \phi, \mathcal{Q}, \theta \} \gets \{ \phi, \mathcal{Q}, \theta \} - \eta \nabla \mathcal{L}_{\text{gen}}$
    \State Discriminator opt: $D_{\text{adv}} \gets D_{\text{adv}} - \eta \nabla \mathcal{L}_{\text{D}}(w, \hat{w})$
\EndWhile
\State $\mathcal{T}(\cdot) \gets \{ \text{Flatten}, E_\phi, \mathcal{Q} \},\; \mathcal{T}^{-1}(\cdot) \gets \{ \mathcal{Q}^{-1}, D_\theta, \text{Unflatten} \}$
\Return $\mathcal{T}(\cdot), \mathcal{T}^{-1}(\cdot)$
\end{algorithmic}
\end{algorithm}

\subsection{Behavioral Cloning Policy}
We train an autoregressive behavioral cloning policy \citep{liu2026oat}, `NACPolicy`, utilizing the frozen \ourshort{} tokenizer. The policy is a causal transformer that predicts the discrete token sequence from observations and instructions.

To provide structured generation, \ourshort{} uses offset codebooks. If the tokenizer has \(n_{\text{q}}\) scales, the policy vocabulary size is \(|\mathcal{V}| = n_{\text{q}} \times V_{\text{bins}} + 1\), accounting for a Beginning-Of-Sequence (\texttt{BOS}) token. Token IDs for scale \(s\) are bounded within \([s \cdot V_{\text{bins}}, (s+1) \cdot V_{\text{bins}})\).

During training, the ground truth discrete codes (a list of \(n_{\text{q}}\) tensors) are offset and concatenated into a flat 1D sequence. The layout dictates that all tokens for scale 0 are predicted first, followed sequentially by tokens for scale 1, and so forth:
\begin{equation}
    C_{\text{flat}} = [\text{BOS}, C^{(0)}_1, \dots, C^{(0)}_{L_0}, C^{(1)}_1, \dots, C^{(1)}_{L_1}, \dots]
\end{equation}
We optimize the policy using a standard causal cross-entropy loss \citep{radford2018improving} over this structured sequence, conditioned on features from the observation encoder.

At inference, the policy autoregressively generates tokens using this fixed layout. It partitions them into per-scale segments, recovers code indices via modulo arithmetic, and passes them to the frozen \ourshort{} detokenizer (Appendix~\ref{app:policy_infer}, Algorithm~\ref{algo:our_policy_infer}). We execute the first \(n_{\text{steps}}\) of the predicted chunk in a receding horizon fashion.

\section{Experiments}
We evaluate NAC's reconstruction fidelity, downstream control performance, and compression--latency tradeoff. Our ablations examine the choice of reconstruction loss, adversarial training, and decoder head.

\subsection{Experimental Setup}

\begin{figure}[ht]
    \centering
    \includegraphics[width=\linewidth]{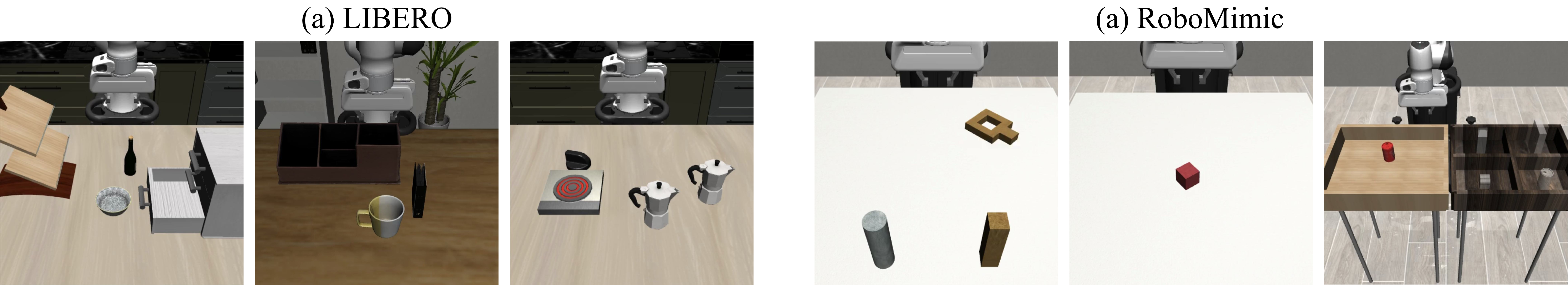}
    \caption{Simulation environments. We benchmark tokenizers across both (a) a LIBERO-10 subset~\citep{liu2023libero} and (b) RoboMimic~\citep{mandlekar2022matters} to assess downstream control performance.}
    \label{fig:sim-envs}
\end{figure}

We compare NAC against continuous-control and discrete-token baselines: Bin \citep{openvla2024}, Diffusion Policy \citep{chi2025diffusion}, FAST \citep{pertsch2025fast}, VQ-VLA \citep{wang2025vqvla}, and OAT \citep{liu2026oat}. These span naive binning, diffusion-based control, hand-designed compression, and learned tokenization. Bin, FAST, VQ-VLA, OAT, and NAC share the same causal transformer policy architecture (\(d=256\), 4 layers, 4 heads), observation encoder, action horizon (32), batch size (256), and training schedule; only the tokenizer and its vocabulary change. Full hyperparameters are in Appendix Tables~\ref{tab:tok_hparams} and~\ref{tab:policy_hparams}. We evaluate on LIBERO-10 \citep{liu2023libero} and RoboMimic \citep{mandlekar2022matters}. Figure~\ref{fig:sim-envs} shows both simulation suites. We report task success rate (\%) for downstream policies and reconstruction MSE for tokenizer ablations. Per-task real-world results are in Appendix Table~\ref{tab:real_world}.

\begin{table*}[t]
\centering
\small
\begin{subtable}[t]{0.48\linewidth}
\centering
\setlength{\tabcolsep}{4pt}
\begin{tabular}[t]{lcc}
\toprule
\textbf{Recon. Loss} & \textbf{Perf. (\%)} & \textbf{MSE} \\
\midrule
L1 & 44.78 \(\pm\) 2.48 & 0.002 \(\pm\) 0.005 \\
MSE & \textbf{49.2 \(\pm\) 1.54} & 0.0008 \(\pm\) 0.0007 \\
DCT & 47.85 \(\pm\) 1.18 & 0.0007 \(\pm\) 0.0008 \\
Log-Mel Spec. & 0 \(\pm\) 0.11 & 0.038 \(\pm\) 0.026 \\
Mel (no log) & 44.6 \(\pm\) 2.0 & 0.0006 \\
Spectrogram & 48.3 \(\pm\) 2.92 & \textbf{0.0002 \(\pm\) 0.001} \\
\bottomrule
\end{tabular}
\caption*{\textbf{(a)} Reconstruction Loss}
\phantomsubcaption
\label{tab:loss_ablation}
\end{subtable}%
\hfill
\begin{subtable}[t]{0.48\linewidth}
\centering
\setlength{\tabcolsep}{4pt}
\begin{tabular}[t]{lcc}
\toprule
\textbf{Discriminator} & \textbf{Perf. (\%)} & \textbf{MSE} \\
\midrule
DAC & 49.45 \(\pm\) 2.02 & 0.0005 \(\pm\) 0.0018 \\
MPD & 46.28 \(\pm\) 1.48 & 0.0005 \(\pm\) 0.0007 \\
MRD & 45.68 \(\pm\) 1.72 & \textbf{0.0004 \(\pm\) 0.0006} \\
None (Spec.) & 0 & 0.35 \(\pm\) 0.12 \\
None (MSE) & \textbf{50.1 \(\pm\) 1.7} & 0.0010 \\
\bottomrule
\end{tabular}
\caption*{\textbf{(b)} Discriminator}
\phantomsubcaption
\label{tab:disc_ablation}
\end{subtable}

\vspace{1.5em}

\begin{subtable}[t]{\linewidth}
\centering
\begin{tabular}[t]{lcc}
\toprule
\textbf{Tokenizer Head} & \textbf{Performance (\%)} & \textbf{MSE} \\
\midrule
ISTFT & \textbf{48.3 \(\pm\) 2.92} & \textbf{0.0002 \(\pm\) 0.001} \\
Linear & 42.1 \(\pm\) 1.54 & 0.0006 \(\pm\) 0.001 \\
\bottomrule
\end{tabular}
\caption*{\textbf{(c)} Decoder Head}
\phantomsubcaption
\label{tab:head_ablation}
\end{subtable}
\caption{Action tokenization ablations on LIBERO-10 \citep{liu2023libero}. Performance (\%) represents downstream policy task success rate; MSE is reconstruction error on 14{,}000 validation action chunks. \textbf{(a)}~Reconstruction loss. \textbf{(b)}~Discriminator; None (Spec.) uses magnitude-only spectrogram loss and None (MSE) uses MSE. \textbf{(c)}~Decoder head.}
\label{tab:combined_ablations}
\end{table*}

\begin{table*}[t]
\centering
\small
\resizebox{\linewidth}{!}{%
\begin{tabular}{lcccccc}
\toprule
\textbf{Environment} & \textbf{Bin \citep{openvla2024}} & \textbf{Diffusion Policy \citep{chi2025diffusion}} & \textbf{FAST \citep{pertsch2025fast}} & \textbf{VQ-VLA \citep{wang2025vqvla}} & \textbf{OAT \citep{liu2026oat}} & \textbf{NAC (Ours)} \\
\midrule
\textbf{LIBERO-10}~\cite{liu2023libero} & 3.95 \(\pm\) 0.8 & 25.48 \(\pm\) 1.3 & 38.02 \(\pm\) 1.3 & 10.85 \(\pm\) 1.85 & 44.17 \(\pm\) 1.2 & \textbf{49.73 \(\pm\) 1.0} \\
\textbf{RoboMimic}~\citep{mandlekar2022matters} & 7.56 \(\pm\) 1.05 & 27.25 \(\pm\) 1.87 & 28.38 \(\pm\) 2.37 & 21.44 \(\pm\) 1.45 & 31.94 \(\pm\) 2.15 & \textbf{33.94 \(\pm\) 1.86} \\
\textbf{Real World} & 6.25 & 22.5 & 40.0 & 31.25 & 40.0 & \textbf{50.0} \\
\bottomrule
\end{tabular}%
}
\caption{Overall manipulation performance (success rate \%) across simulation and real-world environments. Simulation benchmarks were evaluated over 8 seeds with 50 trials per task. Real-world values indicate the average success rate across 8 physical tasks (10 trials each). A full task-by-task breakdown is provided in Appendix Table~\ref{tab:real_world}.}
\label{tab:overall_performance}
\end{table*}

\subsection{Can audio codec architectures model robot actions?}
\label{sec:codec_ablations}
Table~\ref{tab:combined_ablations} shows that neural audio codecs support effective action tokenization with several reconstruction objectives. MSE yields the highest policy success rate, whereas spectrogram loss provides the lowest reconstruction MSE, emphasizing that tokenizers should be evaluated on both metrics.

Log-mel training performs poorly in our configuration. Removing only the logarithm, with the mel scaling and discriminators unchanged, increases success to 44.6\% and restores coarse-codebook usage (Figure~\ref{fig:reconstruction_diagnostics}). This suggests that the relative emphasis on low-energy bands introduced by the logarithm accounts for the failure, rather than the mel scale itself.

The role of adversarial training depends on the reconstruction objective. Magnitude-only spectrogram loss does not explicitly match phase. With flattened actions, sample position determines timestep and channel identity; without a discriminator, reconstructed channels become misaligned and the gripper command is lost (Figure~\ref{fig:reconstruction_diagnostics}), yielding zero success. MSE without a discriminator yields 50.1\% success (Table~\ref{tab:disc_ablation}), comparable to NAC with adversarial training. Adversarial training is therefore not required with the time-domain objective in this setting. The failure under magnitude-only reconstruction also occurs with a linear head, indicating that it is not specific to ISTFT decoding.

\subsection{Do NAC tokens improve performance?}

Having established NAC as a viable tokenizer, we evaluate its downstream control performance on standard simulated manipulation benchmarks \citep{liu2023libero, mandlekar2022matters}. Table~\ref{tab:overall_performance} shows that NAC achieves the highest performance on both LIBERO-10 \citep{liu2023libero} and RoboMimic \citep{mandlekar2022matters}.

On LIBERO-10, NAC outperforms FAST by 11.71 points and OAT by 5.56 points. Similar trends emerge on RoboMimic. These results indicate that better action compression alone is insufficient; the tokenizer must preserve the structure required for next-token policy learning. NAC's multi-scale residual quantization provides a superior interface for autoregressive control compared to hand-designed compression (FAST) and alternative learned tokenizers.

\subsection{Do NAC tokens transfer to the real world?}

\begin{figure}[ht]
    \centering
    \includegraphics[width=0.98\linewidth]{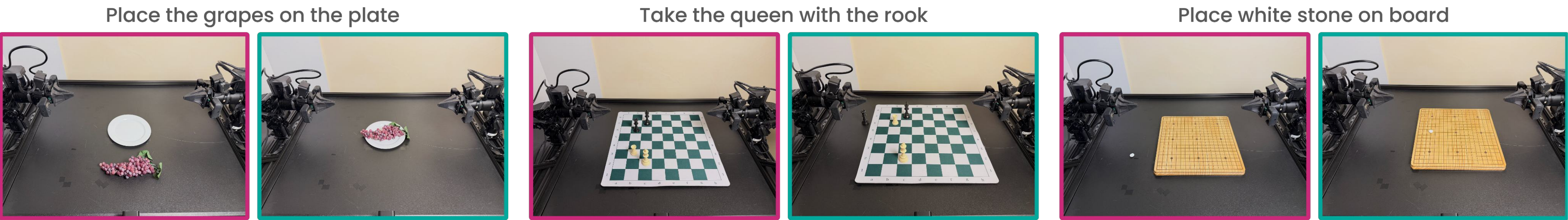}
    \caption{Real-world evaluation tasks. Outlines indicate initial start states (red) and successfully completed states (green).}
    \label{fig:real-world}
    \vspace{-4mm}
\end{figure}

We evaluated all methods on 8 physical manipulation tasks spanning fine grasping, object placement, and deformable control. Figure~\ref{fig:real-world} shows representative tasks; Appendix~\ref{app:real_world} gives full layouts and per-task results. Table~\ref{tab:overall_performance} illustrates that NAC achieves the highest average real-world success rate.

NAC reaches 50\% average success, compared with 40\% for both OAT and FAST. It performs best on grasping grapes and stacking two blocks, while OAT performs best on stacking three blocks and chess. NAC therefore achieves the highest average success.

\subsection{Does stronger compression make fast policies practical?}

\begin{table}[t]
\centering
\scriptsize
\setlength{\tabcolsep}{4pt}
\resizebox{\columnwidth}{!}{
\begin{tabular}{lccccccc}
\toprule
\textbf{Method} & \textbf{Params (M)} & \textbf{Tokens} & \textbf{\(|K|\)} & \textbf{\# of Bits} & \textbf{Enc (ms)} & \textbf{Dec (ms)} & \textbf{Recon (ms)} \\
\midrule
Bin \citep{openvla2024} & 0.002 & 224 & 1024 & 2240 & 0.045 & 0.039 & 0.079 \\
OAT \citep{liu2026oat} & 65.207 & 12 & 1024 & 120 & 0.931 & 2.392 & 3.347 \\
VQ-VLA \citep{wang2025vqvla} & 65.556 & 12 & 1024 & 120 & 7.086 & 4.045 & 11.049 \\
FAST \citep{pertsch2025fast} & 0.000 & 36 & 1024 & 360 & 0.170 & 0.110 & 0.290 \\
NAC (Ours) & 63.006 & 12 & 1024 & 120 & 1.270 & 2.183 & 3.536 \\
\bottomrule
\end{tabular}
}
\caption{Compression and latency statistics for different tokenizers. Recon denotes total reconstruction time on a single Nvidia RTX 4090 GPU.}
\label{tab:latency}
\end{table}

Action tokenization fundamentally aims to reduce the sequence length processed by the policy. Shorter sequences lower autoregressive rollout costs, which is critical as control frequency increases. We compare compression statistics and tokenizer runtime in Table~\ref{tab:latency}.

NAC compresses each action chunk to 12 tokens, matching top learned tokenizers while reducing the token count by nearly 19\(\times\) relative to Bin and 3\(\times\) relative to FAST. Although slower than hand-designed tokenizers like FAST, NAC is significantly faster than VQ-VLA and operates within a practical range for real-time control. This tradeoff provides a highly favorable balance between compression, fidelity, and downstream policy performance, making high-frequency autoregressive policies more plausible.

\section{Discussion}
We introduced the Neural Action Codec (NAC), a multi-scale RVQGAN architecture adapted for robotic action tokenization. Neural audio architectures provide a practical foundation for action tokenization through adaptations to the signal representation, compression rate, and reconstruction objective. NAC provides a compact, structured action space for causal autoregressive policies, yielding higher average success than the evaluated baselines in simulation and across real-world tasks.

\textbf{Limitations.} The flattened 1D action representation requires the sequence length \(H_{\text{a}} \cdot D_{\text{a}}\) to be strictly divisible by the network's combined downsampling ratios (\(K = \prod(\text{ratios}) \times \prod(\text{vq\_scales})\)). Additionally, detokenization requires explicit knowledge of the target action dimensionality to reshape the decoded 1D signal correctly. Flattening couples temporal and channel structure; channel-wise and attention-based encoders remain alternatives. The sensitivity to the inherited RVQ settings and loss weights remains to be studied.

\textbf{Future Work.} We plan to extend NAC to a shared token space across embodiments, train on larger and more diverse action datasets, and apply it to high-frequency control.

\section*{Acknowledgments}

This work was supported in part by the National Science Foundation (NSF) under Grant No. 2520553.

%% file: content/appendix.tex
\section{Method Details}
\label{app:method}

\subsection{Autoregressive Policy Inference}
\label{app:policy_infer}

Algorithm~\ref{algo:our_policy_infer} summarizes inference with the frozen NAC detokenizer. Conditioned on observations and the task, the causal transformer policy generates a fixed number of tokens determined by the tokenizer's per-scale lengths, starting from \texttt{BOS}. After removing \texttt{BOS}, we partition the sequence into per-scale segments and recover code indices modulo \(V_{\text{bins}}\). The detokenizer reconstructs an action chunk, and we execute its first \(n_{\text{steps}}\) steps before collecting new observations and predicting the next chunk.

\begin{algorithm}[H]
\caption{Autoregressive \ourshort{} Policy Inference}
\label{algo:our_policy_infer}
\begin{algorithmic}[1]
\Require Observations $o_{\text{img}}$ from $N_{\text{cam}}$ cameras; task condition $l$; autoregressive policy $\pi(\cdot)$; detokenizer $\mathcal{T}^{-1}$; maximum tokens $L_{\text{max}}$; number of scales $n_{\text{q}}$; bins per scale $V_{\text{bins}}$.
\State Extract features: $f \gets \text{ObsEncoder}(o_{\text{img}}, l)$
\State Initialize sequence: $C_{\text{flat}} \gets [\text{BOS}]$
\While{$|C_{\text{flat}}| \leq L_{\text{max}}$}
    \State Sample next token: $c_{\text{next}} \sim \pi(\,\cdot \mid f, C_{\text{flat}}\,) $
    \State Append token: $C_{\text{flat}} \gets C_{\text{flat}} \oplus c_{\text{next}}$
\EndWhile
\State Remove \texttt{BOS} token from $C_{\text{flat}}$
\State Initialize list for decoded scales: $C_{\text{scales}} \gets \varnothing$
\State Partition $C_{\text{flat}}$ into $n_{\text{q}}$ segments based on the known lengths $\{L_0, \dots, L_{n_{\text{q}}-1}\}$
\For{each segment $S_s$ corresponding to scale $s$}
    \State Recover code indices: $C^{(s)} \gets S_s \pmod{V_{\text{bins}}}$
    \State Append to list: $C_{\text{scales}} \gets C_{\text{scales}} \oplus C^{(s)}$
\EndFor
\State Detokenize to action chunk: $\hat{a}_{1:H_{\text{a}}} \gets \mathcal{T}^{-1}(C_{\text{scales}})$
\Return $\hat{a}_{1:n_{\text{steps}}}$ \Comment{Execute actions in a receding horizon}
\end{algorithmic}
\end{algorithm}

\section{Experimental Details}
\label{app:experiments}

\subsection{Training Hyperparameters}
\label{app:hparams}

Tables~\ref{tab:tok_hparams} and~\ref{tab:policy_hparams} summarize settings from our training configs for LIBERO-10 and RoboMimic. All tokenizer and policy experiments use action horizon 32 and 500 training demonstrations. Dashes indicate settings that do not apply (e.g., non-learned tokenizers).

\begin{table}[ht]
\centering
\small
\setlength{\tabcolsep}{4.5pt}
\renewcommand{\arraystretch}{1.2}
\begin{tabular}{@{}p{0.15\linewidth}p{0.15\linewidth}p{0.14\linewidth}p{0.14\linewidth}p{0.15\linewidth}p{0.15\linewidth}@{}}
\toprule
\textbf{Setting} & \textbf{NAC} & \textbf{Bin} & \textbf{FAST} & \textbf{VQ-VLA} & \textbf{OAT} \\
\midrule
Training steps & 50k & --- & --- & 50k & 50k \\
Discretization & 2-scale RVQ & per-dim.\ bins & DCT+BPE & 12-group VQ & FSQ + registers \\
Codebook size & $1024$ & 1024 & 1024 & $1024 \times 12$ & $1024$ \\
Architecture & SEANet + 12-layer Vocos & --- & pretrained FAST & causal VAE & encoder $L{=}3$, decoder $L{=}10$ \\
\addlinespace
Learning rate & $2{\times}10^{-4}$ & --- & --- & $5{\times}10^{-5}$ & $2{\times}10^{-4}$ \\
LR schedule & cosine & --- & --- & --- & --- \\
\addlinespace
Loss / other & MSE + DAC; $\lambda_{\text{rec}}{=}45$, $\lambda_{\text{commit}}{=}10^3$, $\lambda_{\text{adv}}{=}1$ & $[-1,1]$ & DCT scale 10 & VQ wt.\ 5 & 12 registers \\
\bottomrule
\end{tabular}
\caption{Tokenizer hyperparameters for benchmarked methods. NAC's loss weights follow the audio codec recipe and are not tuned for actions.}
\label{tab:tok_hparams}
\end{table}

\vspace{4mm}

\begin{table}[ht]
\centering
\small
\setlength{\tabcolsep}{8pt}
\renewcommand{\arraystretch}{1.2}
\begin{tabular}{@{}p{0.52\linewidth}r@{}}
\toprule
\textbf{Setting} & \textbf{Value} \\
\midrule
Action horizon & 32 \\
Executed action steps & 16 \\
Observation history length & 2 \\
Training steps & 50k \\
Batch size & 256 \\
\addlinespace
Transformer dimension ($d$) & 256 \\
Transformer layers & 4 \\
Attention heads & 4 \\
Dropout & 0.1 \\
\addlinespace
Policy learning rate & $5{\times}10^{-5}$ \\
Vision encoder learning rate & $1{\times}10^{-5}$ \\
\midrule
\multicolumn{2}{l}{\textit{Autoregressive policies (Bin, FAST, VQ-VLA, OAT, NAC)}} \\
\addlinespace
Top-$k$ sampling & 10 \\
Sampling temperature & 1.0 \\
\midrule
\multicolumn{2}{l}{\textit{Diffusion Policy}} \\
\addlinespace
Inference sampler & DDIM \\
Inference steps & 10 \\
Training timesteps & 100 \\
\bottomrule
\end{tabular}
\caption{Policy hyperparameters. Bin, FAST, VQ-VLA, OAT, and NAC share the same causal transformer policy architecture, observation encoder, and training schedule; only the tokenizer and its vocabulary change. Diffusion-specific settings are listed separately.}
\label{tab:policy_hparams}
\end{table}

\subsection{Reconstruction Diagnostics}
\label{app:reconstruction_diagnostics}

\begin{figure}[ht]
    \centering
    \includegraphics[width=\linewidth]{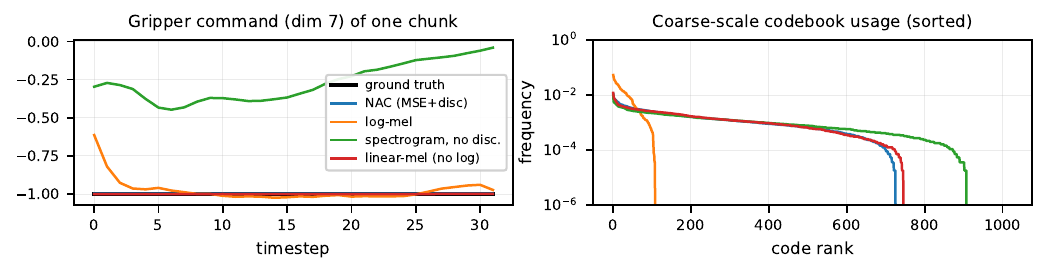}
    \caption{Reconstruction diagnostics on held-out action chunks. \textbf{Left:} Gripper command over one chunk. \textbf{Right:} Sorted coarse-codebook usage over 13{,}757 chunks. Log-mel training uses 108 of 1024 codes; removing the logarithm increases usage to 745, compared with 725 for NAC and 907 for spectrogram training without a discriminator.}
    \label{fig:reconstruction_diagnostics}
\end{figure}

Figure~\ref{fig:reconstruction_diagnostics} shows gripper reconstruction and coarse-codebook usage for the loss ablations in Section~\ref{sec:codec_ablations}. The model trained with magnitude-only spectrogram loss and no discriminator uses 907 coarse codes despite failing to reconstruct the gripper command. A linear map from its quantized features also recovers the actions more accurately than the decoder, indicating that the reconstruction failure does not arise from codebook collapse.

\subsection{Real-World Evaluation}
\label{app:real_world}

\begin{figure}[ht]
    \centering
    \includegraphics[width=0.75\linewidth]{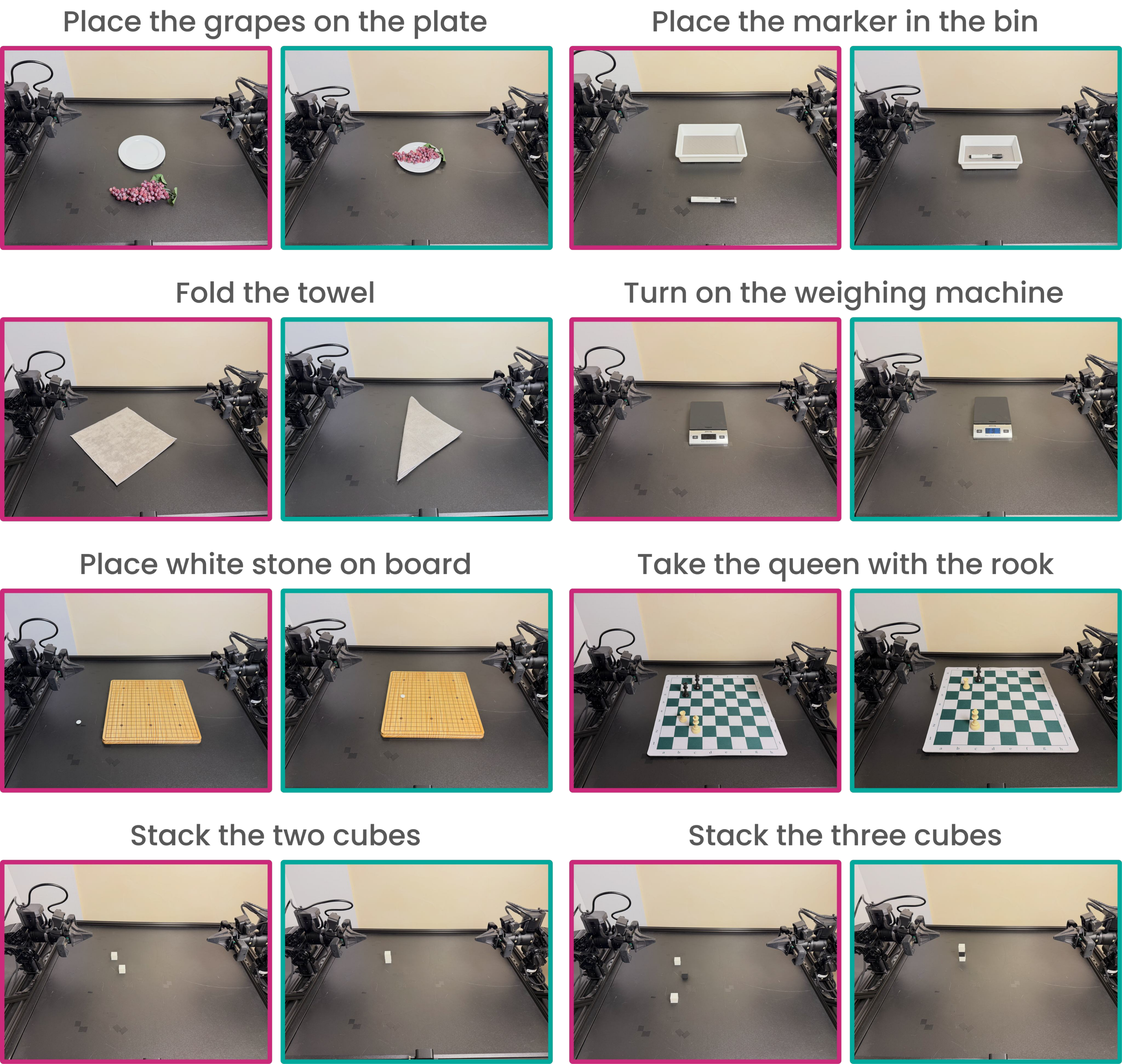}
    \caption{Full Real-world evaluation tasks. Outlines indicate the initial start states (red) and successfully completed states (green)}
    \label{fig:full-real}
\end{figure}

We evaluate all policies trained with these tokenizers on eight physical manipulation tasks with 10 trials per task per method. To keep conditions consistent across policies, we recreate each scene using physical markers outlined with black tape, matching the layouts used during data collection (Figure~\ref{fig:full-real} shows all eight tasks).

\begin{table}[ht]
\centering
\begin{tabular}{lcccccc}
\toprule
\textbf{Task} & \textbf{Bin} & \textbf{Diffusion} & \textbf{FAST} & \textbf{VQ-VLA} & \textbf{OAT} & \textbf{NAC (Ours)} \\
\midrule
Weighing & 50 & 40 & 80 & 90 & 40 & \textbf{90} \\
Grapes & 0 & 30 & 80 & 30 & 80 & \textbf{100} \\
Marker & 0 & 30 & \textbf{60} & 30 & 50 & 50 \\
Two Blocks & 0 & 0 & 0 & 0 & 0 & \textbf{30} \\
Three Blocks & 0 & 0 & 0 & 0 & \textbf{10} & 0 \\
Chess & 0 & 0 & 10 & 0 & \textbf{40} & 10 \\
Place Stone & 0 & 0 & 10 & \textbf{50} & 30 & 40 \\
Fold Towel & 0 & \textbf{80} & \textbf{80} & 50 & 70 & \textbf{80} \\
\midrule
Average & 6.25 & 22.5 & 40 & 31.25 & 40 & \textbf{50} \\
\bottomrule
\end{tabular}
\caption{Real-world manipulation performance reported (success rate \%) across 8 tasks with 10 trials per task.}
\label{tab:real_world}
\end{table}

\begin{table}[ht]
    \centering
    \small
    \setlength{\tabcolsep}{6pt}
    \renewcommand{\arraystretch}{1.2}
    \begin{adjustbox}{max width=\linewidth}
\begin{tabular}{@{}l p{0.28\linewidth} p{0.28\linewidth} p{0.28\linewidth}@{}}
    \toprule
\textbf{Task} & \textbf{Task Prompt} & \textbf{Success Criterion} & \textbf{Failure Criterion} \\
\midrule
Weighing & Turn on the weighing machine. & Scale display on after the arm retracts. & Button pressed but screen not active. \\
Grapes & Place the grapes on the plate. & Grapes transferred onto the plate. & Fewer than half remain on the plate. \\
Marker & Place the marker in the bin. & Marker fully inside the bin at trial end. & Marker outside the bin or dropped outside the workspace. \\
Two Blocks & Stack the two cubes. & Both cubes stacked stably. & Either cube falls, is knocked off the workspace, or the stack is incomplete. \\
Three Blocks & Stack the three cubes. & All three cubes stacked stably. & Any cube falls, is knocked off the workspace, or the stack is incomplete. \\
Chess & Take the queen with the rook. & Rook captures the queen as intended. & Queen not moved off the playing mat, or any piece knocked over. \\
Place Stone & Place white stone on board. & Stone on the target intersection from training. & Stone on any other intersection (illegal move). \\
Fold Towel & Fold the towel. & Towel matches the folded target from training. & Towel not folded, unfolded, or knocked off the workspace. \\
\bottomrule
\end{tabular}
\end{adjustbox}
\caption{Real-world evaluation tasks and trial criteria.}
\label{tab:real_world_tasks}
\end{table}

Each trial runs for up to 1000 action timesteps. A trial ends when the policy succeeds, enters a task-specific failure state, or hits this horizon. An operator may also terminate a trial early if continued execution appears likely to damage the robot or the environment. Success and failure are judged manually using the criteria in Table~\ref{tab:real_world_tasks}.

%% file: main.bib
@inproceedings{openvla2024,
  title     = {{OpenVLA}: An Open-Source Vision-Language-Action Model},
  author    = {Kim, Moo Jin and Pertsch, Karl and Karamcheti, Siddharth and Xiao, Ted and Balakrishna, Ashwin and Nair, Suraj and Rafailov, Rafael and Foster, Ethan P. and Sanketi, Pannag R. and Vuong, Quan and Kollar, Thomas and Burchfiel, Benjamin and Tedrake, Russ and Sadigh, Dorsa and Levine, Sergey and Liang, Percy and Finn, Chelsea},
  booktitle = {8th Annual Conference on Robot Learning},
  year      = {2024},
  url       = {https://openreview.net/forum?id=ZMnD6QZAE6},
}

@article{black2024pi0,
  title   = {$\pi_0$: A Vision-Language-Action Flow Model for General Robot Control},
  author  = {Black, Kevin and Brown, Noah and Driess, Danny and Esmail, Adnan and Equi, Michael and Finn, Chelsea and Fusai, Niccolo and Groom, Lachy and Hausman, Karol and Ichter, Brian and others},
  journal = {arXiv preprint arXiv:2410.24164},
  year    = {2024},
}

@inproceedings{black2025pi0.5,
  title     = {$\pi_{0.5}$: A Vision-Language-Action Model with Open-World Generalization},
  author    = {Black, Kevin and Brown, Noah and Darpinian, James and Dhabalia, Karan and Driess, Danny and Esmail, Adnan and Equi, Michael Robert and Finn, Chelsea and Fusai, Niccolo and Galliker, Manuel Y. and Ghosh, Dibya and Groom, Lachy and Hausman, Karol and Ichter, Brian and Jakubczak, Szymon and Jones, Tim and Ke, Liyiming and LeBlanc, Devin and Levine, Sergey and Li-Bell, Adrian and Mothukuri, Mohith and Nair, Suraj and Pertsch, Karl and Ren, Allen Z. and Shi, Lucy Xiaoyang and Smith, Laura and Springenberg, Jost Tobias and Stachowicz, Kyle and Tanner, James and Vuong, Quan and Walke, Homer and Walling, Anna and Wang, Haohuan and Yu, Lili and Zhilinsky, Ury},
  booktitle = {9th Annual Conference on Robot Learning},
  year      = {2025},
  url       = {https://openreview.net/forum?id=vlhoswksBO},
}

@misc{intelligence2026pi07steerablegeneralistrobotic,
      title={${\pi}_{0.7}$: a Steerable Generalist Robotic Foundation Model with Emergent Capabilities}, 
      author={Physical Intelligence and Bo Ai and Ali Amin and Raichelle Aniceto and Ashwin Balakrishna and Greg Balke and Kevin Black and George Bokinsky and Shihao Cao and Thomas Charbonnier and Vedant Choudhary and Foster Collins and Ken Conley and Grace Connors and James Darpinian and Karan Dhabalia and Maitrayee Dhaka and Jared DiCarlo and Danny Driess and Michael Equi and Adnan Esmail and Yunhao Fang and Chelsea Finn and Catherine Glossop and Thomas Godden and Ivan Goryachev and Lachlan Groom and Haroun Habeeb and Hunter Hancock and Karol Hausman and Gashon Hussein and Victor Hwang and Brian Ichter and Connor Jacobsen and Szymon Jakubczak and Rowan Jen and Tim Jones and Gregg Kammerer and Ben Katz and Liyiming Ke and Mairbek Khadikov and Chandra Kuchi and Marinda Lamb and Devin LeBlanc and Brendon LeCount and Sergey Levine and Xinyu Li and Adrian Li-Bell and Vladislav Lialin and Zhonglin Liang and Wallace Lim and Yao Lu and Enyu Luo and Vishnu Mano and Nandan Marwaha and Aikys Mongush and Liam Murphy and Suraj Nair and Tyler Patterson and Karl Pertsch and Allen Z. Ren and Gavin Schelske and Charvi Sharma and Baifeng Shi and Lucy Xiaoyang Shi and Laura Smith and Jost Tobias Springenberg and Kyle Stachowicz and Will Stoeckle and Jiaming Tang and Jimmy Tanner and Shalom Tekeste and Marcel Torne and Kyle Vedder and Quan Vuong and Anna Walling and Haohuan Wang and Jason Wang and XuDong Wang and Chris Whalen and Samuel Whitmore and Blake Williams and Charles Xu and Sukwon Yoo and Lili Yu and Wuming Zhang and Zhuoyang Zhang and Ury Zhilinsky},
      year={2026},
      eprint={2604.15483},
      archivePrefix={arXiv},
      primaryClass={cs.LG},
      url={https://arxiv.org/abs/2604.15483}, 
}

@article{pertsch2025fast,
  title   = {{FAST}: Efficient Action Tokenization for Vision-Language-Action Models},
  author  = {Pertsch, Karl and Stachowicz, Kyle and Ichter, Brian and Driess, Danny and Nair, Suraj and Vuong, Quan and Mees, Oier and Finn, Chelsea and Levine, Sergey},
  journal = {arXiv preprint arXiv:2501.09747},
  year    = {2025},
}

@article{lee2025molmoact,
  title   = {{MolmoAct}: Action Reasoning Models that can Reason in Space},
  author  = {Lee, Jason and Duan, Jiafei and Fang, Haoquan and Deng, Yuquan and Liu, Shuo and Li, Boyang and Fang, Bohan and Zhang, Jieyu and Wang, Yi Ru and Lee, Sangho and others},
  journal = {arXiv preprint arXiv:2508.07917},
  year    = {2025},
}

@article{zeghidour2021soundstream,
  title     = {{SoundStream}: An End-to-End Neural Audio Codec},
  author    = {Zeghidour, Neil and Luebs, Alejandro and Omran, Ahmed and Skoglund, Jan and Tagliasacchi, Marco},
  journal   = {IEEE/ACM Transactions on Audio, Speech, and Language Processing},
  volume    = {30},
  pages     = {495--507},
  year      = {2021},
  publisher = {IEEE},
}

@article{defossez2023encodec,
  title   = {High Fidelity Neural Audio Compression},
  author  = {D{\'e}fossez, Alexandre and Copet, Jade and Synnaeve, Gabriel and Adi, Yossi},
  journal = {Transactions on Machine Learning Research},
  issn    = {2835-8856},
  year    = {2023},
  url     = {https://openreview.net/forum?id=ivCd8z8zR2},
  note    = {arXiv:2210.13438},
}

@inproceedings{kumar2023dac,
  title     = {High-Fidelity Audio Compression with Improved {RVQGAN}},
  author    = {Kumar, Rithesh and Seetharaman, Prem and Luebs, Alejandro and Kumar, Ishaan and Kumar, Kundan},
  booktitle = {Advances in Neural Information Processing Systems},
  volume    = {36},
  pages     = {27980--27993},
  year      = {2023},
  note      = {arXiv:2306.06546},
}

@inproceedings{siuzdak2024snac,
  title     = {{SNAC}: Multi-Scale Neural Audio Codec},
  author    = {Siuzdak, Hubert and Gr{\"o}tschla, Florian and Lanzend{\"o}rfer, Luca A.},
  booktitle = {Audio Imagination: NeurIPS 2024 Workshop on AI-Driven Speech, Music, and Sound Generation},
  year      = {2024},
  url       = {https://openreview.net/forum?id=PFBF5ctj4X},
}

@article{siuzdak2023vocos,
  title   = {{Vocos}: Closing the Gap Between Time-Domain and Fourier-Based Neural Vocoders for High-Quality Audio Synthesis},
  author  = {Siuzdak, Hubert},
  journal = {arXiv preprint arXiv:2306.00814},
  year    = {2023},
}

@article{borsos2023audiolm,
  title     = {{AudioLM}: A Language Modeling Approach to Audio Generation},
  author    = {Borsos, Zal{\'a}n and Marinier, Rapha{\"e}l and Vincent, Damien and Kharitonov, Eugene and Pietquin, Olivier and Sharifi, Matt and Roblek, Dominik and Teboul, Olivier and Grangier, David and Tagliasacchi, Marco and others},
  journal   = {IEEE/ACM Transactions on Audio, Speech, and Language Processing},
  volume    = {31},
  pages     = {2523--2533},
  year      = {2023},
  publisher = {IEEE},
}

@article{wang2023valle,
  title   = {Neural Codec Language Models are Zero-Shot Text to Speech Synthesizers},
  author  = {Wang, Chengyi and Chen, Sanyuan and Wu, Yu and Zhang, Ziqiang and Zhou, Long and Liu, Shujie and Chen, Zhuo and Liu, Yanqing and Wang, Huaming and Li, Jinyu and others},
  journal = {arXiv preprint arXiv:2301.02111},
  year    = {2023},
}

@article{yang2023uniaudio,
  title   = {{UniAudio}: An Audio Foundation Model Toward Universal Audio Generation},
  author  = {Yang, Dongchao and Tian, Jinchuan and Tan, Xu and Huang, Rongjie and Liu, Songxiang and Chang, Xuankai and Shi, Jiatong and Zhao, Sheng and Bian, Jiang and Zhao, Zhou and others},
  journal = {arXiv preprint arXiv:2310.00704},
  year    = {2023},
}

@article{dong2026actioncodec,
  title   = {{ActionCodec}: What Makes for Good Action Tokenizers},
  author  = {Dong, Zibin and Liu, Yicheng and Zhang, Shiduo and Ye, Baijun and Yuan, Yifu and Ni, Fei and Gong, Jingjing and Qiu, Xipeng and Zhao, Hang and Li, Yinchuan and others},
  journal = {arXiv preprint arXiv:2602.15397},
  year    = {2026},
}

@inproceedings{liu2026faster,
  title     = {{FAST}er: Toward Powerful and Efficient Autoregressive Vision--Language--Action Models with Learnable Action Tokenizer and Block-wise Decoding},
  author    = {Liu, Yicheng and Zhang, Shiduo and Dong, Zibin and Ye, Baijun and Yuan, Tianyuan and Yu, Xiaopeng and Yin, Linqi and Lu, Chenhao and Shi, Junhao and Yu, Luca Jiang-Tao and Zheng, Liangtao and Gong, Jingjing and Jiang, Tao and Qiu, Xipeng and Zhao, Hang},
  booktitle = {The Fourteenth International Conference on Learning Representations},
  year      = {2026},
  url       = {https://openreview.net/forum?id=k6nTUFoqeT},
  note      = {arXiv:2512.04952},
}

@article{liu2026oat,
  title   = {{OAT}: Ordered Action Tokenization},
  author  = {Liu, Chaoqi and Han, Xiaoshen and Gao, Jiawei and Zhao, Yue and Chen, Haonan and Du, Yilun},
  journal = {arXiv preprint arXiv:2602.04215},
  year    = {2026},
}

@inproceedings{wang2025vqvla,
  title     = {{VQ-VLA}: Improving Vision-Language-Action Models via Scaling Vector-Quantized Action Tokenizers},
  author    = {Wang, Yating and Zhu, Haoyi and Liu, Mingyu and Yang, Jiange and Fang, Hao-Shu and He, Tong},
  booktitle = {Proceedings of the IEEE/CVF International Conference on Computer Vision},
  year      = {2025},
  note      = {arXiv:2507.01016},
}

@inproceedings{kaneko2022istftnet,
  title     = {{iSTFTNet}: Fast and Lightweight Mel-Spectrogram Vocoder Incorporating Inverse Short-Time Fourier Transform},
  author    = {Kaneko, Takuhiro and Tanaka, Kou and Kameoka, Hirokazu and Seki, Shogo},
  booktitle = {ICASSP 2022--2022 IEEE International Conference on Acoustics, Speech and Signal Processing (ICASSP)},
  pages     = {6207--6211},
  year      = {2022},
  organization = {IEEE},
}

@article{jang2021univnet,
  title   = {{UnivNet}: A Neural Vocoder with Multi-Resolution Spectrogram Discriminators for High-Fidelity Waveform Generation},
  author  = {Jang, Won and Lim, Dan and Yoon, Jaesam and Kim, Bongwan and Kim, Juntae},
  journal = {arXiv preprint arXiv:2106.07889},
  year    = {2021},
}

@inproceedings{kong2020hifi,
  title     = {{HiFi-GAN}: Generative Adversarial Networks for Efficient and High Fidelity Speech Synthesis},
  author    = {Kong, Jungil and Kim, Jaehyeon and Bae, Jaekyoung},
  booktitle = {Advances in Neural Information Processing Systems},
  volume    = {33},
  pages     = {17022--17033},
  year      = {2020},
}

@inproceedings{lee2022autoregressive,
  title     = {Autoregressive Image Generation using Residual Quantization},
  author    = {Lee, Doyup and Kim, Chiheon and Kim, Saehoon and Cho, Minsu and Han, Wook-Shin},
  booktitle = {Proceedings of the IEEE/CVF Conference on Computer Vision and Pattern Recognition},
  pages     = {11523--11532},
  year      = {2022},
}

@article{chi2025diffusion,
  title     = {Diffusion Policy: Visuomotor Policy Learning via Action Diffusion},
  author    = {Chi, Cheng and Xu, Zhenjia and Feng, Siyuan and Cousineau, Eric and Du, Yilun and Burchfiel, Benjamin and Tedrake, Russ and Song, Shuran},
  journal   = {The International Journal of Robotics Research},
  volume    = {44},
  number    = {10-11},
  pages     = {1684--1704},
  year      = {2025},
  publisher = {Sage Publications},
  note      = {arXiv:2303.04137},
}

@article{goodfellow2014generative,
  title={Generative adversarial nets},
  author={Goodfellow, Ian J and Pouget-Abadie, Jean and Mirza, Mehdi and Xu, Bing and Warde-Farley, David and Ozair, Sherjil and Courville, Aaron and Bengio, Yoshua},
  journal={Advances in neural information processing systems},
  volume={27},
  year={2014}
}

@article{yang2025qwen3,
  title={Qwen3 technical report},
  author={Yang, An and Li, Anfeng and Yang, Baosong and Zhang, Beichen and Hui, Binyuan and Zheng, Bo and Yu, Bowen and Gao, Chang and Huang, Chengen and Lv, Chenxu and others},
  journal={arXiv preprint arXiv:2505.09388},
  year={2025}
}

@inproceedings{
cadene2026lerobot,
title={LeRobot:  An Open-Source Library for End-to-End Robot Learning},
author={Remi Cadene and Simon Alibert and Francesco Capuano and Michel Aractingi and Adil Zouitine and Pepijn Kooijmans and Jade Choghari and Martino Russi and Caroline Pascal and Steven Palma and Dana Aubakirova and Mustafa Shukor and Jess Moss and Alexander Soare and Quentin Lhoest and Quentin Gallou{\'e}dec and Thomas Wolf},
booktitle={The Fourteenth International Conference on Learning Representations},
year={2026},
url={https://openreview.net/forum?id=CiZMMAFQR3}
}

@misc{beyer2024paligemmaversatile3bvlm,
      title={PaliGemma: A versatile 3B VLM for transfer}, 
      author={Lucas Beyer and Andreas Steiner and André Susano Pinto and Alexander Kolesnikov and Xiao Wang and Daniel Salz and Maxim Neumann and Ibrahim Alabdulmohsin and Michael Tschannen and Emanuele Bugliarello and Thomas Unterthiner and Daniel Keysers and Skanda Koppula and Fangyu Liu and Adam Grycner and Alexey Gritsenko and Neil Houlsby and Manoj Kumar and Keran Rong and Julian Eisenschlos and Rishabh Kabra and Matthias Bauer and Matko Bošnjak and Xi Chen and Matthias Minderer and Paul Voigtlaender and Ioana Bica and Ivana Balazevic and Joan Puigcerver and Pinelopi Papalampidi and Olivier Henaff and Xi Xiong and Radu Soricut and Jeremiah Harmsen and Xiaohua Zhai},
      year={2024},
      eprint={2407.07726},
      archivePrefix={arXiv},
      primaryClass={cs.CV},
      url={https://arxiv.org/abs/2407.07726}, 
}

@misc{touvron2023llama2openfoundation,
      title={Llama 2: Open Foundation and Fine-Tuned Chat Models}, 
      author={Hugo Touvron and Louis Martin and Kevin Stone and Peter Albert and Amjad Almahairi and Yasmine Babaei and Nikolay Bashlykov and Soumya Batra and Prajjwal Bhargava and Shruti Bhosale and Dan Bikel and Lukas Blecher and Cristian Canton Ferrer and Moya Chen and Guillem Cucurull and David Esiobu and Jude Fernandes and Jeremy Fu and Wenyin Fu and Brian Fuller and Cynthia Gao and Vedanuj Goswami and Naman Goyal and Anthony Hartshorn and Saghar Hosseini and Rui Hou and Hakan Inan and Marcin Kardas and Viktor Kerkez and Madian Khabsa and Isabel Kloumann and Artem Korenev and Punit Singh Koura and Marie-Anne Lachaux and Thibaut Lavril and Jenya Lee and Diana Liskovich and Yinghai Lu and Yuning Mao and Xavier Martinet and Todor Mihaylov and Pushkar Mishra and Igor Molybog and Yixin Nie and Andrew Poulton and Jeremy Reizenstein and Rashi Rungta and Kalyan Saladi and Alan Schelten and Ruan Silva and Eric Michael Smith and Ranjan Subramanian and Xiaoqing Ellen Tan and Binh Tang and Ross Taylor and Adina Williams and Jian Xiang Kuan and Puxin Xu and Zheng Yan and Iliyan Zarov and Yuchen Zhang and Angela Fan and Melanie Kambadur and Sharan Narang and Aurelien Rodriguez and Robert Stojnic and Sergey Edunov and Thomas Scialom},
      year={2023},
      eprint={2307.09288},
      archivePrefix={arXiv},
      primaryClass={cs.CL},
      url={https://arxiv.org/abs/2307.09288}, 
}

@misc{fang2026molmoact2actionreasoningmodels,
      title={MolmoAct2: Action Reasoning Models for Real-world Deployment}, 
      author={Haoquan Fang and Jiafei Duan and Donovan Clay and Sam Wang and Shuo Liu and Weikai Huang and Xiang Fan and Wei-Chuan Tsai and Shirui Chen and Yi Ru Wang and Shanli Xing and Jaemin Cho and Jae Sung Park and Ainaz Eftekhar and Peter Sushko and Karen Farley and Angad Wadhwa and Cole Harrison and Winson Han and Ying-Chun Lee and Eli VanderBilt and Rose Hendrix and Suveen Ellawela and Lucas Ngoo and Joyce Chai and Zhongzheng Ren and Ali Farhadi and Dieter Fox and Ranjay Krishna},
      year={2026},
      eprint={2605.02881},
      archivePrefix={arXiv},
      primaryClass={cs.RO},
      url={https://arxiv.org/abs/2605.02881}, 
}

@inproceedings{darcet2024vision,
  title={Vision Transformers Need Registers},
  author={Darcet, Timoth{\'e}e and Oquab, Maxime and Mairal, Julien and Bojanowski, Piotr},
  booktitle={International Conference on Learning Representations (ICLR)},
  year={2024}
}

@inproceedings{rippel2014learning,
  title={Learning ordered representations with nested dropout},
  author={Rippel, Oren and Gelbart, Michael and Adams, Ryan},
  booktitle={International Conference on Machine Learning},
  pages={1746--1754},
  year={2014},
  organization={PMLR}
}

@article{vaswani2017attention,
  title={Attention is all you need},
  author={Vaswani, Ashish and Shazeer, Noam and Parmar, Niki and Uszkoreit, Jakob and Jones, Llion and Gomez, Aidan N and Kaiser, {\L}ukasz and Polosukhin, Illia},
  journal={Advances in neural information processing systems},
  volume={30},
  year={2017}
}

@article{ahmed1974discrete,
  title={Discrete cosine transform},
  author={Ahmed, Nasir and Natarajan, T\_ and Rao, Kamisetty R},
  journal={IEEE transactions on Computers},
  volume={100},
  number={1},
  pages={90--93},
  year={1974},
  publisher={IEEE}
}

@article{zhao2016loss,
  title={Loss functions for image restoration with neural networks},
  author={Zhao, Hang and Gallo, Orazio and Frosio, Iuri and Kautz, Jan},
  journal={IEEE Transactions on computational imaging},
  volume={3},
  number={1},
  pages={47--57},
  year={2016},
  publisher={IEEE}
}

@article{zhao2023learning,
  title={Learning fine-grained bimanual manipulation with low-cost hardware},
  author={Zhao, Tony Z and Kumar, Vikash and Levine, Sergey and Finn, Chelsea},
  journal={arXiv preprint arXiv:2304.13705},
  year={2023}
}

@inproceedings{chen2021decision,
  title={Decision transformer: reinforcement learning via sequence modeling},
  author={Chen, Lili and Lu, Kevin and Rajeswaran, Aravind and Lee, Kimin and Grover, Aditya and Laskin, Michael and Abbeel, Pieter and Srinivas, Aravind and Mordatch, Igor},
  booktitle={Proceedings of the 35th International Conference on Neural Information Processing Systems},
  pages={15084--15097},
  year={2021}
}

@inproceedings{torabi2018behavioral,
  title={Behavioral cloning from observation},
  author={Torabi, Faraz and Warnell, Garrett and Stone, Peter},
  booktitle={Proceedings of the 27th International Joint Conference on Artificial Intelligence},
  pages={4950--4957},
  year={2018}
}

@inproceedings{tagliasacchi20_interspeech,
  title     = {{SEANet: A Multi-Modal Speech Enhancement Network}},
  author    = {Marco Tagliasacchi and Yunpeng Li and Karolis Misiunas and Dominik Roblek},
  year      = {2020},
  booktitle = {{Interspeech 2020}},
  pages     = {1126--1130},
  doi       = {10.21437/Interspeech.2020-1563},
  issn      = {2958-1796},
}

@article{clevert2015fast,
  title={Fast and accurate deep network learning by exponential linear units (elus)},
  author={Clevert, Djork-Arn{\'e} and Unterthiner, Thomas and Hochreiter, Sepp},
  journal={arXiv preprint arXiv:1511.07289},
  volume={4},
  number={5},
  pages={11},
  year={2015}
}

@article{salimans2016weight,
  title={Weight normalization: A simple reparameterization to accelerate training of deep neural networks},
  author={Salimans, Tim and Kingma, Durk P},
  journal={Advances in neural information processing systems},
  volume={29},
  year={2016}
}

@inproceedings{he2016deep,
  title={Deep residual learning for image recognition},
  author={He, Kaiming and Zhang, Xiangyu and Ren, Shaoqing and Sun, Jian},
  booktitle={Proceedings of the IEEE conference on computer vision and pattern recognition},
  pages={770--778},
  year={2016}
}

@inproceedings{isola2017image,
  title={Image-to-image translation with conditional adversarial networks},
  author={Isola, Phillip and Zhu, Jun-Yan and Zhou, Tinghui and Efros, Alexei A},
  booktitle={Proceedings of the IEEE conference on computer vision and pattern recognition},
  pages={1125--1134},
  year={2017}
}

@article{gray1984vector,
  title={Vector quantization},
  author={Gray, Robert},
  journal={IEEE Assp Magazine},
  volume={1},
  number={2},
  pages={4--29},
  year={1984},
  publisher={IEEE}
}

@inproceedings{sennrich2016neural,
  title={Neural machine translation of rare words with subword units},
  author={Sennrich, Rico and Haddow, Barry and Birch, Alexandra},
  booktitle={Proceedings of the 54th annual meeting of the association for computational linguistics (volume 1: long papers)},
  pages={1715--1725},
  year={2016}
}

@inproceedings{macqueen1967multivariate,
  title={Multivariate observations},
  author={MacQueen, J},
  booktitle={Proceedings ofthe 5th Berkeley symposium on mathematical statisticsand probability},
  volume={1},
  pages={281--297},
  year={1967},
  organization={University of California press Oakland, CA, USA}
}

@article{van2017neural,
  title={Neural discrete representation learning},
  author={Van Den Oord, Aaron and Vinyals, Oriol and others},
  journal={Advances in neural information processing systems},
  volume={30},
  year={2017}
}

@article{roy2018theory,
  title={Theory and experiments on vector quantized autoencoders},
  author={Roy, Aurko and Vaswani, Ashish and Neelakantan, Arvind and Parmar, Niki},
  journal={arXiv preprint arXiv:1805.11063},
  year={2018}
}

@article{targ2016resnet,
  title={Resnet in resnet: Generalizing residual architectures},
  author={Targ, Sasha and Almeida, Diogo and Lyman, Kevin},
  journal={arXiv preprint arXiv:1603.08029},
  year={2016}
}

@inproceedings{liu2022convnet,
  title={A convnet for the 2020s},
  author={Liu, Zhuang and Mao, Hanzi and Wu, Chao-Yuan and Feichtenhofer, Christoph and Darrell, Trevor and Xie, Saining},
  booktitle={Proceedings of the IEEE/CVF conference on computer vision and pattern recognition},
  pages={11976--11986},
  year={2022}
}

@article{radford2018improving,
  title={Improving language understanding by generative pre-training},
  author={Radford, Alec and Narasimhan, Karthik and Salimans, Tim and Sutskever, Ilya and others},
  year={2018},
  journal={Openai},
  publisher={San Francisco, CA, USA}
}

@article{liu2023libero,
  title={Libero: Benchmarking knowledge transfer for lifelong robot learning},
  author={Liu, Bo and Zhu, Yifeng and Gao, Chongkai and Feng, Yihao and Liu, Qiang and Zhu, Yuke and Stone, Peter},
  journal={Advances in Neural Information Processing Systems},
  volume={36},
  pages={44776--44791},
  year={2023}
}

@inproceedings{mandlekar2022matters,
  title={What Matters in Learning from Offline Human Demonstrations for Robot Manipulation},
  author={Mandlekar, Ajay and Xu, Danfei and Wong, Josiah and Nasiriany, Soroush and Wang, Chen and Kulkarni, Rohun and Fei-Fei, Li and Savarese, Silvio and Zhu, Yuke and Mart{\'\i}n-Mart{\'\i}n, Roberto},
  booktitle={Conference on Robot Learning},
  pages={1678--1690},
  year={2022},
  organization={PMLR}
}

@article{kiranyaz20211d,
  title={1D convolutional neural networks and applications: A survey},
  author={Kiranyaz, Serkan and Avci, Onur and Abdeljaber, Osama and Ince, Turker and Gabbouj, Moncef and Inman, Daniel J},
  journal={Mechanical systems and signal processing},
  volume={151},
  pages={107398},
  year={2021},
  publisher={Elsevier}
}
